%% file: ACMEASE26.tex
\documentclass[sigconf,authorversion,nonacm]{acmart}

\acmConference[EASE 2026]{The 30th International Conference on Evaluation and Assessment in Software Engineering}{9–12 June, 2026}{Glasgow, Scotland, United Kingdom}
\AtBeginDocument{%
  }

\usepackage{changepage}
\usepackage{framed}
\usepackage{tabularx}
\usepackage{subcaption}
\usepackage{balance}

\usepackage[acronym,nohypertypes={acronym}]{glossaries}
\makeatletter
 {\par\unskip\endMakeFramed}
\makeatother

\definecolor{formalshade}{rgb}{0.93,0.93,0.93}
\definecolor{darkblue}{rgb}{0.2, 0.2, 0.2}

\newenvironment{formal}{%
  \def\FrameCommand{%
    \hspace{1pt}%
    {\color{darkblue}\vrule width 2pt}%
    {\color{formalshade}\vrule width 4pt}%
    \colorbox{formalshade}%
  }%
  \MakeFramed{\advance\hsize-\width\FrameRestore}%
  \noindent\hspace{-1pt}
  \begin{adjustwidth}{}{7pt}%
  \vspace{2pt}\vspace{2pt}%
}
{%
  \vspace{3pt}\end{adjustwidth}\endMakeFramed%
}

\newcounter{resultcounter}
\newenvironment{result}{\begin{formal}
    \refstepcounter{resultcounter}
    \noindent \hspace{-4pt}
}{
\end{formal}
}

\usepackage{hyperref,url}
\usepackage{xspace}
\newcommand{\etal}[1]{#1~\textit{et al.}}
\newcommand\ie{\emph{i.e.},\xspace}
\newcommand\eg{\emph{e.g.},\xspace}

\newcommand{\citesec}[1]{Section~\ref{sec:#1}}
\newcommand{\citefig}[1]{Figure~\ref{fig:#1}}

\newcommand{\citetable}[1]{Table~\ref{table:#1}}

\definecolor{brickred}{rgb}{0.8, 0.25, 0.33}
\definecolor{upforestgreen}{rgb}{0.0, 0.27, 0.13}
\newacronym{spl}{SPL}{{\emph{Software Product Line}}}
\newacronym{emf}{EMF}{{\emph {Eclipse Modeling Framework}}}
\newacronym{sple}{SPLE}{\emph{Software Product Line Engineering}}
\newacronym{xml}{XML}{\emph{eXtensible Markup Language}}
\newacronym{cot}{COT}{\emph{Chain of Thought}}
\newacronym{rag}{RAG}{\emph{Retrieval Augmented Generation}}
\newacronym{cdf}{CDF}{\emph{Cumulative Distribution Function}}
\newacronym{vllm}{vLLM}{\emph{virtual Large Language Model}}
\newacronym{sftm}{SFTM}{\emph{Similarity-based Flexible Tree Matching}}
\newacronym{dom}{DOM}{\emph{Document Object Model}}
\newacronym{uvl}{UVL}{\emph{Universal Variability Language}}
\newacronym{coc}{CoC}{\emph{Chain of Code}}
\newacronym{cove}{CoVe}{\emph{Chain of verification}}
\newacronym{aafm}{AAFM}{\emph{Automated Analysis of Feature Models}}
\newacronym{fca}{FCA}{\emph{Formal Concept Analysis}}
\newacronym{eas}{EAs}{\emph{Evolutionary Algorithms}}
\newacronym{cao}{CaO}{\emph{Clone-and-Own}}
\newacronym{bleu}{BLEU}{\emph{BiLingual Evaluation Understudy}}
\newacronym{dtr}{DTR}{\emph{Decision Tree Regression}}
\newacronym{rfr}{RFR}{\emph{Random Forest Regression}}
\newacronym{mlp}{MLP}{\emph{Multilayer Perceptron Neural Network}}
\newacronym{mae}{MAE}{\emph{Mean Absolute Error}}
\newacronym{mse}{MSE}{\emph{Mean Square Error}}
\newacronym{mape}{MAPE}{\emph{Mean Absolute Percentage Error}}
\newacronym{r2}{R²}{\emph{Coefficient of Determination}}
\newacronym{rmse}{RMSE}{\emph{Root Mean Square Error}}
\newacronym{fosd}{FOSD}{\emph{Feature-oriented software development }}
\newacronym{ai}{AI}{\emph{Artificial Intelligence}}
\newacronym{shap}{SHAP}{\emph{Shapley Additive Explanations}}
\newacronym{yasa}{YASA}{\emph{Yet Another Sampling Algorithm}}
\newacronym{rapl}{RAPL}{\emph{Running Average Power Limit}}
\newacronym{tgi}{TGI}{\emph{Text Generation Inference}}
\newcommand{\best}[1]{\textbf{#1}\xspace}

\newacronym[
    \glslongpluralkey={\emph{Large Language Models}}, 
    \glsshortpluralkey={LLMs}
]{llm}{LLM}{\emph{Large Language Model}}
\newacronym[
    \glslongpluralkey={\emph{Software Product Lines}}, 
    \glsshortpluralkey={SPLs}
]{spls}{SPL}{\emph{Software Product Line}}
\newacronym[
    \glslongpluralkey={\emph{Feature Models}}, 
    \glsshortpluralkey={FMs}
]{fm}{FM}{\emph{Feature Model}}

\begin{document}

\title{Pimp~My~LLM: Leveraging Variability~Modeling to~Tune Inference~Hyperparameters}

\author{Nada Zine}
 \affiliation{
   \institution{Univ. Lille, CNRS, Inria, Centrale Lille, UMR 9189 CRIStAL, F-59000, Lille}
   \city{Lille}
   \country{France}}
 \email{nada.zine@inria.fr}

\author{Clément Quinton}
 \affiliation{
   \institution{Univ. Lille, CNRS, Inria, Centrale Lille, UMR 9189 CRIStAL, F-59000, Lille}
   \city{Lille}
   \country{France}}
 \email{clement.quinton@univ-lille.fr}

\author{Romain Rouvoy}
 \affiliation{
   \institution{Univ. Lille, CNRS, Inria, Centrale Lille, UMR 9189 CRIStAL, F-59000, Lille}
   \city{Lille}
   \country{France}}
 \email{romain.rouvoy@univ-lille.fr}

\begin{abstract}
\glspl{llm} are being increasingly used across a wide range of tasks.
However, their substantial computational demands raise concerns about the energy efficiency and sustainability of both training and inference. Inference, in particular, dominates total compute usage, making its optimization crucial. Recent research has explored optimization techniques and analyzed how configuration choices influence energy consumption.
Yet, the vast configuration space of inference servers makes exhaustive empirical evaluation infeasible due to combinatorial explosion.

In this paper, we introduce a new perspective on this problem by treating \glspl{llm} as configurable systems and applying variability management techniques to systematically analyze inference-time configuration choices. 
We evaluate our approach on the Hugging Face Transformers library by representing generation hyperparameters and their constraints using a feature-based variability model, sampling representative configurations, measuring their energy consumption, latency, accuracy, and learning predictive models from the collected data.
Our results show that variability modeling effectively manages the complexity of \gls{llm} inference configurations.
It enables systematic analysis of hyperparameters effects and interactions, reveals trade-offs, and supports prediction of inference behavior from a limited number of measurements.

Overall, this work opens a new research direction that bridges software engineering and machine learning by leveraging variability modeling for the efficient and sustainable configuration of \glspl{llm}.
\end{abstract}


\maketitle

\glsresetall
\input{chapters/introduction}
\input{chapters/related-work}
\input{chapters/approach}
\input{chapters/experiment-setup}
\input{chapters/results}
\input{chapters/threats-to-validity}
\input{chapters/conclusion}
\section*{Data Availability}
All code, notebooks, and supplementary information regarding hyperparameter choices and experimental setup are publicly available, along with the data supporting the findings of this study, at \url{https://doi.org/10.5281/zenodo.17375044}.
\input{chapters/acknowledgments}
\bibliographystyle{ACM-Reference-Format}
\balance
\bibliography{bibfile.bib}

\end{document}

%% file: chapters/introduction.tex
\section{Introduction}
\label{sec:introduction}
\glspl{llm} have transformed both research and industrial practices~\cite{zhao2023survey}, particularly in software engineering, where they now assist with code generation, bug fixing, documentation, and numerous other tasks~\cite{zheng2025towards}. 
To support their growing adoption and further improve performance, major efforts have been invested in scaling computational infrastructure and deploying optimized inference systems~\cite{stojkovic2025dynamollm}.
However, this growth raises concerns about energy efficiency and environmental impact, as \glspl{llm} and other \gls{ai} workloads are highly resource-intensive~\cite{zhao2022green}.

Recent research has sought to better characterize and mitigate the environmental footprint of \glspl{llm} during its training and inference phases.
In particular, improving inference efficiency is essential for reducing costs and carbon emissions~\cite{zhou2024survey}, as \glspl{llm} are executed millions of times per day at a large scale, with inference accounting for over $90$\% of total \gls{llm} compute cycles~\cite{fu2025llmco2,patel2024characterizing}.
Inference efficiency has thus become a central research focus, and researchers have investigated various inference parameters and their impact on diverse metrics related to energy consumption, performance, and accuracy (\eg~\cite{samsi2023words,luccioni2024power,fernandez2025energy,martinez2025impact}).
Some studies have focused more specifically on generation hyperparameters---\ie parameters that control generation behavior such as beam width, temperature, top-p or caching strategy~\cite{shi2402thorough,arias2025decoding,nik2025impact}.
However, inference servers such as Hugging Face Transformers~\cite{wolf2019huggingface}, \gls{vllm}~\cite{kwon2023efficient}, and \gls{tgi}\footnote{\url{https://github.com/huggingface/text-generation-inference}}
exhibit a wide range of such parameters~\cite{park2025survey}. The possible combinations of these parameters result in an enormous configuration space, where interactions between options are often non-linear or implicit. 
This results in two main challenges: \emph{(i)} configuring inference systems manually in practice is challenging, as many parameter configurations are invalid, redundant, or interdependent; and \emph{(ii)} exhaustive empirical evaluation is computationally infeasible.
Consequently, most recent studies focus primarily on a limited set of commonly used parameters and analyze them in isolation, without systematically exploring or modeling parameter combinations. 

These limitations mirror a long-standing challenge in software engineering: reasoning about highly configurable systems. 
Such systems are characterized by built-in variability, offering a set of configuration options that can be combined in different ways to adapt their functional behavior and performance to diverse user requirements~\cite{alves2020sampling,agh2024software}. 
To manage this variability, various techniques have been developed for modeling and reasoning about large configuration spaces~\cite{chen2009variability,czarnecki2012cool,schaefer2012software}.
Among these, \emph{feature modeling} has emerged as a foundational formalism~\cite{czarnecki2007feature,batory2005feature,marques2019software,berger2013survey}. 
\glspl{fm} describes configurable systems through hierarchical features, dependencies, and constraints, and can be translated into propositional logic for automated reasoning~\cite{galindo2019automated}. 
While feature modeling has been successfully applied in domains such as cloud configuration and performance optimization, it has not, to the best of our knowledge, been used to explore the configuration space of \glspl{llm}. Instead, recent work has primarily explored the opposite direction, using \glspl{llm} to assist with variability-related tasks such as variability implementation~\cite{acher2023programming,galindo2023large}, software re-engineering~\cite{acher2023generative,stumpfle2024automating,stumpfle2025large}, co-evolution of variability models and artifacts~\cite{kebaili2024empirical,zine2025llm}, and configuration-related bug classification~\cite{lopez2025configuration}.

In this paper, we consider Hugging Face Transformers as a highly configurable system. 
In particular, we focus on the inference phase and its generation hyperparameters.
Our goal is threefold: \emph{(i)} to ease configuration by grouping all relevant generation hyperparameters and explicitly modeling their constraints; \emph{(ii)} to complement existing analyses by studying how generation hyperparameters and their interactions affect energy, latency, accuracy, and the trade-offs among these measurements; and \emph{(iii)} to propose predictive models for unseen configurations.

Our main contributions are as follows: 
\begin{itemize}
    \item \textbf{Feature-based modeling of \gls{llm} inference:} We encode the variability of the Hugging Face Transformers library in a publicly available feature model\footnote{\url{https://doi.org/10.5281/zenodo.17375044}}, and report how such a model supports systematic exploration over its vast inference configuration space.
    \item \textbf{Optimal trade-offs:} Leveraging the feature model and our systematic exploration methodology, we identify configurations that achieve the best trade-off between energy consumption, latency, and accuracy in our case-study.
    \item \textbf{Prediction methodology:} We develop a four-step approach---modeling, sampling, measurement, and learning---to predict energy consumption, latency, and accuracy of unseen configurations.
\end{itemize}
The remainder of the paper is structured as follows. 
In \citesec{related}, we review related work. 
In \citesec{approach}, we present our proposed approach. 
\citesec{setup} describes the methodology used to evaluate the approach, and \citesec{results} reports the obtained results. 
Broader discussion and limitations are provided in \citesec{limitations}. 
Finally, we conclude the paper and outline directions for future work in \citesec{conclusion}.

%% file: chapters/related-work.tex
\section{Related Work}
\label{sec:related}

\subsection{Impact of Generation Hyperparameters on LLM Inference Performance, Energy Consumption \& Accuracy}
Several studies have investigated how the inference performance, energy consumption, and accuracy of \glspl{llm} are influenced by configuration choices. Collectively, these works show that \glspl{llm} inference efficiency depends on a wide range of parameters, from model- and input-level factors~\cite{samsi2023words,luccioni2024power,maliakel2025investigating,wilkins2024offline,fu2025llmco2,martinez2025impact}, system- and workload-level
parameters~\cite{coignion2024green,fernandez2025energy,stojkovic2024towards,husom2025sustainable,wu2025unveiling}, to system- and hardware-level configurations~\cite{patel2024characterizing,lazuka2024llm,caravaca2025prompts}, with several works spanning multiple of these dimensions. Within this body of work, generation hyperparameters are sometimes varied as part of broader experimental setups, but they are rarely the primary object of study. 
More recent efforts have instead focused explicitly on understanding the role of these hyperparameters.
\etal{Shi} provide a thorough empirical evaluation of decoding hyperparameters, examining their effects on generation quality and performance across a wide range of tasks and models~\cite{shi2402thorough}. \etal{Arias} extended this work by analyzing additional hyperparameter combinations across more models and evaluating text quality with multiple metrics~\cite{arias2025decoding}. Finally, \etal{Nik} complemented these analyses by focusing specifically on how generation hyperparameters influence GPU energy consumption alongside output quality across diverse tasks and configurations~\cite{nik2025impact}.
Together, these studies show that \glspl{llm} inference efficiency depends on numerous parameters across different levels. However, existing work typically examines only limited subsets of parameters at each level, without considering interactions or constraints between them. This limitation can make it difficult to configure models or apply recommendations in practice, and may hide insights that could be revealed by studying such interactions. Furthermore, studies focusing particularly on generation hyperparameters still lack a comprehensive analysis of their impact on overall resource consumption, the trade-offs between latency, energy consumption, accuracy, and the development of predictive models for unseen configurations.

\subsection{Application Domains of Variability Modeling}
Variability modeling has been widely applied in numerous domains to manage configurable software systems~\cite{galindo2019automated}.
For instance, variability management has been particularly important for operating systems such as the Linux kernel. 
\etal{Kuiter} analyzed the evolution of its feature models over twenty years to evaluate configurations, compare metrics, and predict the future of the kernel’s configurability~\cite{kuiter2025configurable}. 
In the cloud domain, \etal{Quinton} proposed a platform relying on feature models combined with a domain model to select the most suitable cloud environment~\cite{quinton2016saloon}. 
In robotics, \etal{Brugali} discussed software variability, domain analysis and modeling, and reference architectures for configurable robotic systems~\cite{brugali2020software}.
Similarly, in \emph{Internet of Things} (IoT), \etal{Trujillo} proposed a method to manage variability across multiple SPLs and generate new product portfolios for IoT applications~\cite{trujillo2020multiple}.
Even in areas such as browser fingerprinting, \etal{Huyghe} introduced a tool based on feature models to reason over the relationships between configuration parameters and browser fingerprints~\cite{huyghe2025browserfm,huyghe2025fp}.
Applications of variability modeling have also been extended to machine learning. 
\etal{Camillieri} and \etal{Ghofrani} applied variability management concepts to manage variability in machine learning workflows and deep neural network architectures, respectively~\cite{camillieri2016towards,ghofrani2019applying}.
Finally, \etal{Gomez} explored the use of variability to study the creation of combined ML models~\cite{gomez2024exploring}.

However, despite these advances, the application of variability modeling \emph{to represent the configuration space of \glspl{llm} themselves} remains largely unexplored. 
To the best of our knowledge, our work is the first to explicitly model and analyze \glspl{llm} variability using feature modeling techniques.

\subsection{Learning Software Configuration Spaces}

A major research direction emerging from variability modeling is the \emph{sampling–measuring–learning} pattern. 
This approach consists of learning from a small set of sampled configurations rather than exhaustively exploring the entire configuration space, which is infeasible for highly configurable systems. 
\etal{Pereira} presented a systematic literature review of the different methods proposed for each step of this pattern and highlighted applications such as prediction, optimization, and dynamic configuration~\cite{pereira2021learning}. 
Following this idea, several studies have applied the pattern across different contexts and for diverse applications.
\etal{Guégain} explored the impact of web stack configurations on performance relying on feature models, and developed a predictive model to automate optimized configurations~\cite{guegain2025exploring}. 
\etal{Kumara} modeled the variability of cloud environments to build performance prediction models and leveraged explainable AI to interpret model outputs~\cite{kumara2022focloud}. 
Similarly, \etal{Couto} used the pattern to predict the energy consumption of a single software product~\cite{couto2017products}.
Overall, these studies show the power of learning from configuration samples to predict performance and energy consumption in configurable systems. 
Our work extends these principles to the inference phase of \glspl{llm}, enabling a systematic exploration of a larger configuration space. 

%% file: chapters/approach.tex
\section{Variability-Driven Energy Prediction of LLM~Inference}
\label{sec:approach}
This section presents our approach for systematically exploring the configuration space of \gls{llm} inference by leveraging variability modeling. 
An overview of the workflow is shown in \citefig{overview}.
\begin{figure}[b]
   \centering
   \includegraphics[width=\linewidth]{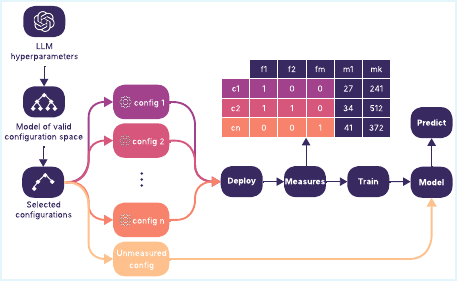}
   \caption{Overview of the proposed approach.}
   \label{fig:overview}
\end{figure}

\subsection{Hugging Face Transformers as a Proof of Concept}
To illustrate our approach, we use the Hugging Face Transformers library~\cite{wolf2019huggingface}, an open-source library providing pretrained models for a wide range of machine learning tasks and support for both training and inference. 
While other inference servers optimized for production, such as \gls{vllm}~\cite{kwon2023efficient}, could also be considered, their primary focus on performance optimization does not align with the main objective of this work. In contrast, Hugging Face Transformers offers two key advantages that are important for our investigation: \emph{(i)} it is well-documented and widely adopted, and \emph{(ii)} it exposes a larger set of configurable hyperparameters, which is essential for exploring the inference configuration space.

\subsection{Modeling Hugging Face Transformers}
To systematically analyze and predict the behavior of \glspl{llm} under different inference configurations, we first need to make their variability explicit. Inference frameworks, such as Hugging Face Transformers, expose a large number of hyperparameters with complex dependencies. Given the combinatorial explosion of possible parameter combinations and their complex interdependencies, manual exploration or ad-hoc tuning is impractical. Moreover, these relationships are often scattered across documentation and source code, making systematic exploration and automated analysis impractical.
 
To address this challenge, we represent \gls{llm} inference variability using \glspl{fm}, the most widely adopted formalism for modeling configurable software systems~\cite{marques2019software,berger2013survey,czarnecki2012cool}.
A \gls{fm} captures variability in terms of \emph{features}, which can be mandatory, optional, or organized into feature groups, and expresses dependencies through \emph{cross-tree constraints} (\eg selecting one feature requires or excludes another).
Importantly, feature models can be translated into propositional formulas, enabling automated reasoning tasks such as enumerating valid configurations, detecting invalid or dead features, and filtering configurations that satisfy given properties~\cite{galindo2019automated}. These capabilities are essential in our context, as the inference configuration space grows exponentially with the number of parameters.
\begin{figure*}[ht]
    \centering
    \includegraphics[width=0.9\linewidth]{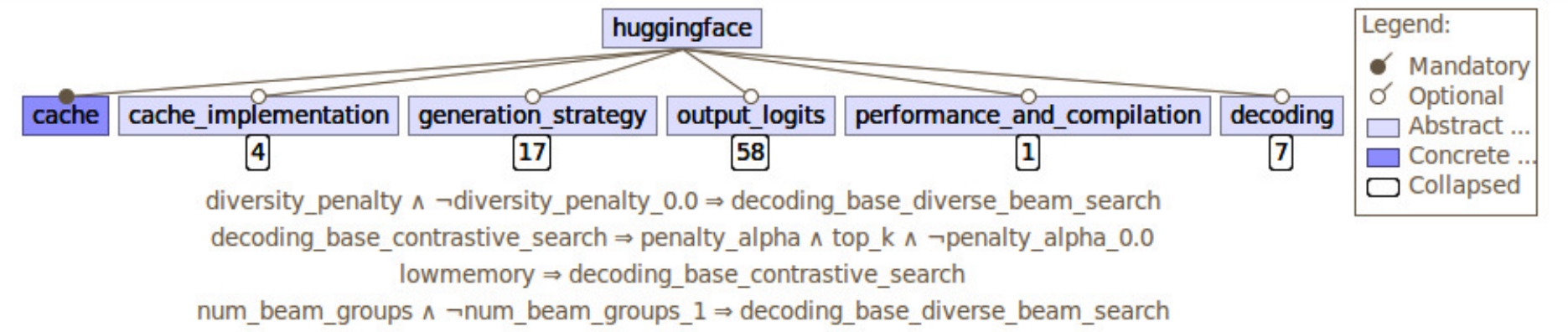}
    \caption{The variability of Hugging Face Transformers explored in this paper. Cross-tree constraints are partially shown.}
    \label{fig:fm}
\end{figure*}

\citefig{fm} presents the resulting feature model for Hugging Face Transformers inference. 
The model contains $96$ features, of which $67$ are concrete, resulting in approximately $9.37 \times 10^{12}$ valid configurations, illustrating the infeasibility of exhaustive exploration without automated reasoning.
Each feature corresponds either to a Boolean parameter or to a discretized value of a numerical parameter. 
For example, \texttt{temperature\_0.7} denotes that the \texttt{temperature} parameter is fixed to $0.7$, and \texttt{do\_sample} enables sampling. 
Dependencies and constraints between features were extracted from official documentation, source code inspection, and empirical testing, ensuring that all configurations generated from the model are valid and executable.



We modeled a large subset of generation hyperparameters exposed by Hugging Face Transformers,\footnote{\url{https://huggingface.co/docs/transformers/main_classes/text_generation}} while excluding or fixing those not relevant to our study. 
For example, parameters such as \texttt{max\_length} and \texttt{stop\_strings} were fixed to prevent truncated or excessively long outputs. 

Because many parameters admit a wide range of values, encoding all possible values would lead to an intractable configuration space. 
We, therefore, discretized parameter domains following recommendations from prior work and from Hugging Face Transformers documentation~\cite{shi2402thorough, arias2025decoding, nik2025impact}, ensuring representative variability while maintaining tractability.
By explicitly modeling Hugging Face Transformers inference variability as a feature model, we make its configuration space analyzable and amenable to automation. 
This model serves as the foundation for the subsequent phases of our approach: systematic sampling of representative configurations, empirical measurement of energy consumption, latency, accuracy, and learning predictive models from the collected data.

\subsection{Sampling LLM Configurations}
Given the exponential size of the configuration space, exhaustively measuring all valid \gls{llm} inference configurations is infeasible. 
Instead, we rely on sampling techniques to select a small yet representative subset of configurations. 
Prior work in the variability community has shown that such samples are sufficient to support meaningful analysis, optimization, and the construction of accurate predictive models~\cite{guegain2021reducing,kumara2022focloud}.
Several sampling strategies have been proposed to address this challenge. 
Among the most widely used are random sampling and interaction-aware (\emph{t-wise}) sampling approaches. 
While random sampling generates valid configurations randomly, \emph{t-wise} sampling explicitly targets interactions between configuration options, ensuring that combinations of up to $t$ features are covered at least once~\cite{al2016tool}. 
This property is particularly important when configuration options interact in non-trivial ways, as is the case for \gls{llm} inference parameters.

In this study, we adopted three complementary sampling strategies: two interaction-aware strategies based on \emph{2-wise} coverage, and one random strategy used as a point of comparison. 
Among the available t-wise sampling algorithms, we chose \emph{YASA}~\cite{krieter2020yasa} and \emph{ICPL}~\cite{johansen2012algorithm}, both of which achieve state-of-the-art coverage efficiency.
For the random strategy, we used the implementation provided by the FeatureIDE platform\footnote{Which we used to encode the feature model depicted in \citefig{fm}.} and set the number of generated configurations equal to the number of features, as recommended in other studies~\cite{guo2018data,kumara2022focloud}. 
\subsection{Measuring LLM Inference Performance}
Each sampled configuration is executed and measured according to three metrics: energy consumption, performance, and accuracy.

\textbf{Energy consumption}
We relied on software-based measurement tools commonly used in recent \gls{llm} studies~\cite{coignion2024green,patel2024characterizing,coignion2025faster,alizadeh2025language,nik2025impact}: \emph{perf} (via the \gls{rapl} interface) for CPU measurements, and \emph{nvidia-smi} for GPU power draw estimation. This approach ensured full control over the measurement process and compatibility with any GPU, facilitating future extensions of this work.

\textbf{Code evaluation}
To evaluate the generated code, we assessed the functional correctness of the generated code.
The functional correctness of an \gls{llm} assesses the extent to which the generated code satisfies the specification provided in the prompt. 
We thus adopted the \emph{pass@k} metric based on the unbiased estimator introduced by \etal{Chen} \cite{chen2021evaluating}, which measures the probability that at least one of $k$ generated samples passes all test cases.
Formally, for $n$ total samples, $c$ correct samples, and $E$ denoting the expectation over all problems, pass@k is defined as:
\[
\text{pass@}k := \mathbb{E}_{\text{Problems}} \left[ 1 - \frac{\binom{n-c}{k}}{\binom{n}{k}} \right]
\]
We used $n = 1$ sample per problem to compute pass@1.

\textbf{Performance}
We evaluated performance by measuring latency, defined here as the time elapsed between submitting a request and receiving its response.

\subsection{Learning \& Validation}
After collecting measurements across all sampled configurations, we use the resulting dataset to train and validate predictive models of \gls{llm} inference behavior.
Our goal is to assess whether models trained on a limited yet representative set of configurations can accurately predict energy consumption and accuracy for unseen configurations.
We relied on \gls{rfr}~\cite{breiman2001random}, a tree-based learning technique that has been widely adopted for learning software configuration spaces~\cite{pereira2021learning}. 
Model selection was performed through grid search combined with $k$-fold cross-validation on the training set. Once the best configuration was selected, the model was retrained on the full training set and evaluated on a test set obtained from a $80$\% / $20$\% train–test split.
To evaluate prediction accuracy, we used two complementary and widely adopted standard metrics~\cite{chicco2021coefficient}:
\begin{itemize}
    \item \textbf{\gls{r2}}: quantifies the proportion of variance in the target metric explained by the model. Values range from $0$ to $1$, with higher values indicating better model performance; however, \gls{r2} can be negative if the model performs worse than a simple baseline predicting the mean.
    \item \textbf{\gls{mae}}: calculates the average absolute difference between predicted and observed values. It provides a linear measure of prediction error in the same units as the target variable. Lower \gls{mae} values indicate higher predictive accuracy.
    \item \textbf{\gls{mape}}: extends \gls{mae} by expressing the error as a percentage of the observed values, effectively normalizing the absolute error relative to the magnitude of the target. Lower MAPE values indicate higher predictive accuracy.
\end{itemize}

%% file: chapters/experiment-setup.tex
\section{Methodology}
\label{sec:setup}
This section describes the methodology employed to evaluate our proposed approach. 
We first present the addressed research questions, followed by a description of the dataset. 
We then introduce the selected \glspl{llm}, outline the experimental setup, and detail the conducted experiments. 

\subsection{Research Questions}

\paragraph{\textbf{RQ1}: \textit{Which individual inference configuration options and feature interactions have the greatest impact on energy consumption, latency, and accuracy, as identified through variability modeling?}}

\paragraph{\textbf{RQ2}: \textit{Which Pareto-optimal trade-offs between energy consumption, performance, and accuracy can be identified?}} 

\paragraph{\textbf{RQ3}: \textit{To what extent can machine learning models accurately predict energy consumption, performance, and accuracy for unseen \gls{llm} inference configurations?}} 

\subsection{Dataset}
To evaluate our approach, we focused on \gls{llm} tasks involving code generation, as they naturally support quantitative evaluation through automated testing. Among existing datasets, \emph{HumanEval}~\cite{chen2021evaluating} is the most widely adopted benchmark for this purpose.\footnote{\url{https://huggingface.co/datasets?other=code-generation}} It contains $164$ hand-written Python programming problems, each associated with reference implementations and unit tests, which make it well-suited for evaluating functional correctness via the pass@k metric.
For this study, we used \emph{HumanEval+}~\cite{liu2023your}, an extended version of \emph{HumanEval} that provides more rigorous and comprehensive test cases.
We used the original problem prompts without applying any additional prompt engineering or task-specific prompting strategies.

\subsection{Selected LLMs}
We selected \glspl{llm} from the \emph{BigCodeBench} leaderboard,\footnote{\url{https://bigcode-bench.github.io/}} a benchmark designed to evaluate \glspl{llm} on its true programming capabilities~\cite{zhuo2024bigcodebench}. 
We applied three selection criteria: \textit{(i)} open-source availability, \textit{(ii)} model size under 8B parameters, and \textit{(iii)} full access to the decoding hyperparameters evaluated in this study.  
We focused on open-source \glspl{llm} to ensure accessibility and facilitate reproducible experimentation. 
We also limited our selection to models with fewer than 8B parameters, as experiments with larger models would have been slower and more resource-intensive with Hugging Face Transformers, without offering additional value to this study.
From the leaderboard, we selected the two highest-performing models that satisfied these criteria at the time of our experiments.
To enable comparative analysis, we included one additional model: the smaller counterpart of the first model.
The selected models are listed in~\citetable{llmsStudied}. 
For clarity, throughout the remainder of the paper, we refer to each model by its family name and size only, \eg \emph{Qwen-7B} for \emph{Qwen2.5-Coder-7B-Instruct}.
\begin{table}[h]
    \caption{\glspl{llm} considered in our study. Size is in billions of parameters.}
    \label{table:llmsStudied}
    \centering
    \renewcommand{\arraystretch}{1}
    \begin{tabular}{l l r}
        \hline
        \textbf{LLM Model} & \textbf{Model family} & \textbf{Sizes} \\
        \hline
        OpenCoder-8B-Instruct & infly & 8\,B \\
        Qwen2.5-Coder-7B-Instruct & Qwen & 7.62\,B \\
        Qwen2.5-Coder-3B-Instruct & Qwen & 3.09\,B \\
        \hline
    \end{tabular}
\end{table}

\subsection{Experimental Setup}
By following the approach described above, we evaluated each of the $254$ configurations ($77$ YASA, $81$ ICPL, and $96$ RANDOM) produced by our sampling algorithms using the $164$ prompts of \emph{HumanEval+} and our three selected \glspl{llm}. 
Each configuration was executed ten times, resulting in $7{,}620$ runs and $1{,}249{,}680$ prompts. 
To mitigate potential hardware-induced bias, the $7{,}620$ runs were executed in a fully randomized order and distributed across our machines. This setup ensures that neither the different configurations nor their ten repetitions were performed in sequence.
To run our experiments with Hugging Face Transformers, we used \texttt{device\_map="auto"} to load the model across our GPUs. 
All models were run with \texttt{bfloat16} precision, a fixed batch size of $32$, and a maximum token of $512$.

Before evaluating each configuration, we performed an internal \emph{calibration phase} of $30$ seconds~\cite{coignion2025faster}. 
This calibration serves two purposes: 
\emph{(i)} adjusting the measurement coefficient that converts raw hardware counters into actual energy units, ensuring that subsequent energy measurements are accurate; 
and \emph{(ii)} capturing the idle consumption, during which the system remains at rest, providing a baseline energy level and allowing a short cool-down before launching the next configuration. 
Across all experiments, idle consumption measurements exhibited a coefficient of variation of $0.02$, indicating high measurement stability. 
In addition to this internal calibration, we introduced a \emph{warm-up phase} before measuring energy consumption for GPU workloads. This phase mitigates cold-start effects such as memory allocation, CUDA graph capture, and just-in-time (JIT) kernel compilation~\cite{fernandez2025energy}. Without this warm-up, initial iterations may include overheads, potentially biasing energy and performance measurements. 

Each of the $7{,}620$ runs was thus preceded by an individual warm-up phase consisting of six iterations. These iterations used the same hyperparameters as the target configuration, along with the experiment's batch size and the longest prompts, to simulate a worst-case inference scenario. The choice of six iterations was determined through a preliminary benchmark, in which we empirically measured the warm-up effect on the same setup. Although the system showed relatively stable behavior early on, we maintained six iterations to ensure energy measurements stabilized within a 3\% threshold over three consecutive runs. In the reported results, the energy consumption of each configuration corresponds to the execution of all $164$ prompts, averaged over ten repetitions.
We report the absolute energy consumption---\ie total power usage including background processes.
Nevertheless, the idle power, together with the energy measured during the evaluation of each configuration, can be used to compute the marginal energy consumption of the samples---\ie the additional energy solely caused by the execution of the workload, excluding baseline consumption~\cite{jagannadharao2024beginner}.

All experiments were conducted on a node from the Grid5000 platform,\footnote{\url{https://www.grid5000.fr/}} equipped with an AMD EPYC 7513 (Zen 3) processor, 512\,GiB of memory, and $4$ Nvidia A100-SXM4-40GB (40\,GiB) GPUs. 
The system ran Debian~11 (Bullseye) with the Linux 5.10.0-28-amd64 kernel.

%% file: chapters/results.tex
\section{Results}
\label{sec:results}
This section presents the experimental results addressing the three research questions introduced in this study.

\paragraph*{\textbf{RQ1}: \textit{Which individual inference configuration options and feature interactions have the greatest impact on energy consumption, latency, and accuracy, as identified through variability modeling?}}

\citefig{variations-cdf} shows variations in energy consumption and accuracy across all configurations generated by all sampling methods for each selected \glspl{llm}.
\emph{Qwen-3B} and \emph{OpenCoder-8B} align closely in terms of energy consumption and latency for approximately 85\% of configurations, indicating similar efficiency. After 40\%, \emph{OpenCoder-8B} starts diverging slightly from Qwen models, consuming more resources. Beyond 80\%, the gap widens: Qwen models remain tight, while \emph{OpenCoder-8B} shifts sharply to the right, indicating higher energy costs and latency. Regarding accuracy (Pass@1), the distributions are similar up to 60\% of configurations. At the upper end, \emph{Qwen-3B} and \emph{OpenCoder-8B} reach the same maximum accuracy, with \emph{Qwen-7B} surpassing them by a small margin.

\begin{figure}[t!]
    \centering
    \includegraphics[width=0.95\linewidth]{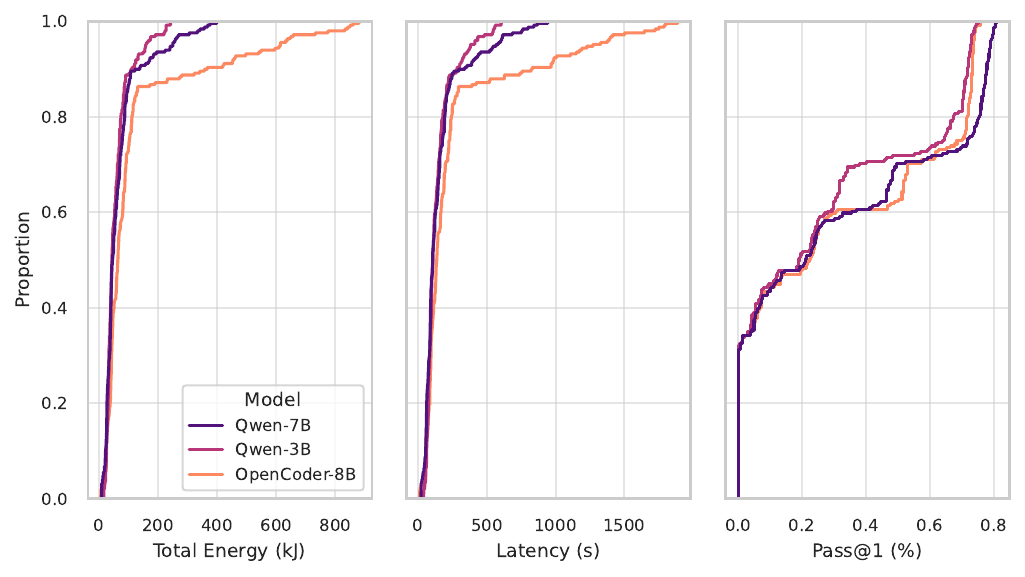}
    \caption{\emph{Cumulative Distribution Function} (CDF) of energy consumption, performance, and accuracy across different configurations for each evaluated \gls{llm}.}
    \label{fig:variations-cdf}
\end{figure}

\textbf{\textit{Feature-wise Analysis}}.
To assess the impact of individual inference hyperparameters on energy consumption and accuracy, we adopt the \emph{feature-wise analysis} proposed by \etal{Guégain}~\cite{guegain2021reducing}.
This method isolates the contribution of each feature to a given metric by aggregating measurements across sampled configurations.
Concretely, a matrix of size $n \times m$ is constructed, where $n$ denotes the number of features and $m$ the configurations. 
For each configuration, the observed metric value is assigned to the columns corresponding to the features it contains, while all other entries remain zero. 
For example, if a configuration \texttt{C} includes features \texttt{F1} and \texttt{F3} and consumes $50$~kJ, the corresponding entries for \texttt{F1} and \texttt{F3} are set to $50$~kJ.
This method provides an aggregated, feature-level view of their impact.

\begin{figure}[h]
  \centering
   \includegraphics[width=\linewidth]{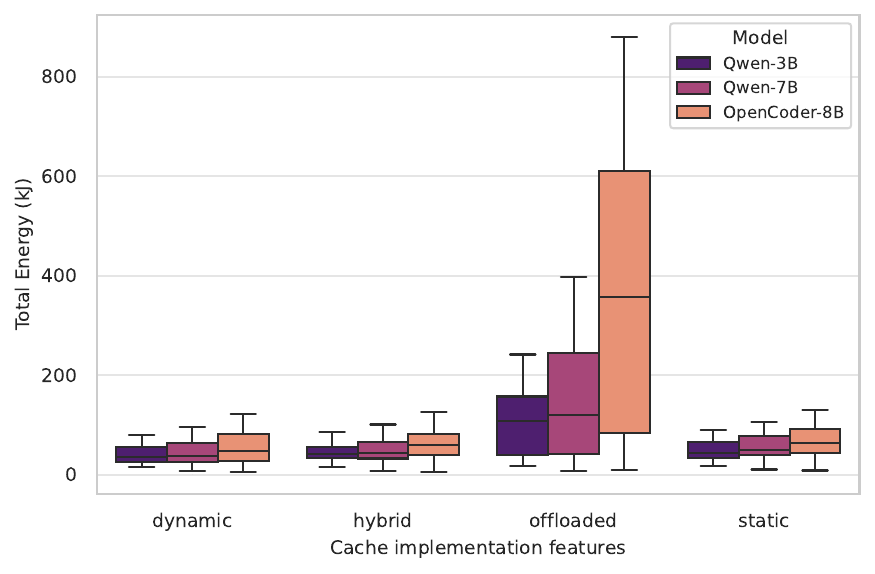}
  \caption{Feature-wise analysis of energy consumption per cache implementation.}
  \label{fig:feature-wise-generation_total_energy_kj_mean_cache}
\end{figure}

\citetable{feature} reports five hyperparameter values with the strongest impact on energy consumption, latency, and accuracy as identified by feature-wise analysis. 
On the energy and latency dimensions, the same features consistently appear as the most influential, showing a strong correlation between energy consumption and latency. 
In both measurements, using an \texttt{offloaded cache} leads to the largest increase in energy consumption. 
The cache implementation feature determines how key–value (KV) pairs are stored and reused during generation. 
Offloading the cache transfers KV data between GPU and CPU memory to reduce GPU memory usage, but this mechanism introduces significant data transfer overhead.
In contrast, alternative cache implementations—such as \texttt{dynamic}, \texttt{hybrid}, and \texttt{static}—exhibit consistently lower energy consumption, as illustrated in Figure~\ref{fig:feature-wise-generation_total_energy_kj_mean_cache}.
These cache strategies exhibit similar trends across all evaluated \glspl{llm}.

Beyond cache implementation, the number of beams and beam groups, both fixed at $4$, are also among the features with the strongest impact on energy and latency. 
The number of beams controls how many parallel hypotheses are explored during generation,
while beam groups further partition the beams into independent groups, which require additional forward passes. 
Both mechanisms increase the number of forward passes per token, leading to higher computational cost, and increased latency and energy consumption.

\begin{table*}[t]
\centering
\caption{Top five features with the strongest impact (absolute value) on energy consumption, latency, and accuracy, identified by the feature-wise analysis.}
\label{table:feature}
\renewcommand{\arraystretch}{1}
\begin{subtable}[t]{0.29\textwidth}
\centering
\caption{Energy consumption ($\Delta$ Energy, in kJ).}
\begin{tabular}{l l r}
\toprule
\textbf{Feature} & \textbf{Value} & $\boldsymbol{\Delta}$ \textbf{Energy} \\
\midrule
cache &     offloaded & +89.48  \\
num-beam-group   & 4 & +58.03 \\
num-beams   & 4 & +50.30 \\
diversity-penalty & 0.4 & +46.81 \\
decoding & diverse & +45.69 \\
\bottomrule
\end{tabular}
\end{subtable}
\hfill
\begin{subtable}[t]{0.29\textwidth}
\centering
\caption{Latency ($\Delta$ Latency, in s).}
\begin{tabular}{l l r}
\toprule
\textbf{Feature} & \textbf{Value} & $\boldsymbol{\Delta}$ \textbf{Latency} \\
\midrule
cache &     offloaded & +233.51 \\
num-beam-group   & 4 & +129.35 \\
num-beams         & 4 & +111.83 \\
diversity-penalty & 0.4 & +100.66 \\
decoding & diverse & +100.37 \\
\bottomrule
\end{tabular}
\end{subtable}
\hfill
\begin{subtable}[t]{0.29\textwidth}
\centering
\caption{Accuracy ($\Delta$ Pass@1).}
\begin{tabular}{l l r}
\toprule
\textbf{Feature} & \textbf{Value} & $\boldsymbol{\Delta}$ \textbf{Pass@1} \\
\midrule
no-repeat-ngram-size & 0         & +0.68 \\
decoding           & greedy & +0.46\\
encoder-repetition           & 1.0 & +0.41\\
repetition-penalty & 1.0 & +0.24\\
low-memory & True & -0.23\\
\bottomrule
\end{tabular}
\end{subtable}
\hfill
\end{table*}

\begin{table*}[t]
\caption{Top-5 feature combinations with greedy search that have the strongest impact (absolute value) on energy consumption and accuracy, identified by the pair-wise analysis.}
\label{table:feature_pair}
\centering
\renewcommand{\arraystretch}{1}
\begin{subtable}[t]{0.30\textwidth}
\centering
\caption{Energy consumption ($\Delta$ Energy, in kJ).}
\begin{tabular}{l l r}
\toprule
\textbf{Feature} & \textbf{Value} & $\boldsymbol{\Delta}$ \textbf{Energy} \\
\midrule
cache & static & +13.76 \\
encoder-repetition-p & 1.3 & +13.76\\
do-sample & True & -8.59\\
top-k  & 5 & +7.04 \\
temperature & 0.3 & -2.2\\
\bottomrule
\end{tabular}
\end{subtable}
\hfill
\begin{subtable}[t]{0.30\textwidth}
\centering
\caption{Latency ($\Delta$ Latency, in s).}
\begin{tabular}{l l r}
\toprule
\textbf{Feature} & \textbf{Value} & $\boldsymbol{\Delta}$ \textbf{Latency} \\
\midrule
cache & static & +48.12 \\
encoder-repetition-p & 1.3 & +48.12\\
do-sample & True & -21.79\\
top-k  & 5 & +20.87 \\
top-p & 0.8 & -7.46\\
\bottomrule
\end{tabular}
\end{subtable}
\hfill
\begin{subtable}[t]{0.30\textwidth}
\centering
\caption{Accuracy ($\Delta$ Pass@1).}
\begin{tabular}{l l r}
\toprule
\textbf{Features} & \textbf{Value} & $\boldsymbol{\Delta}$ \textbf{Pass@1} \\
\midrule
repetition-penalty & 0.6 & -0.65 \\
do-sample & True & +0.37 \\
encoder-repetition-p & 1.3 & -0.36\\
top-p  & 0.8 & +0.23 \\
temperature  & 1.2 & -0.20 \\
\bottomrule
\end{tabular}
\end{subtable}
\end{table*}

\begin{figure*}[t]
  \centering
    \begin{subfigure}{0.44\linewidth}
    \centering
    \includegraphics[width=\linewidth]{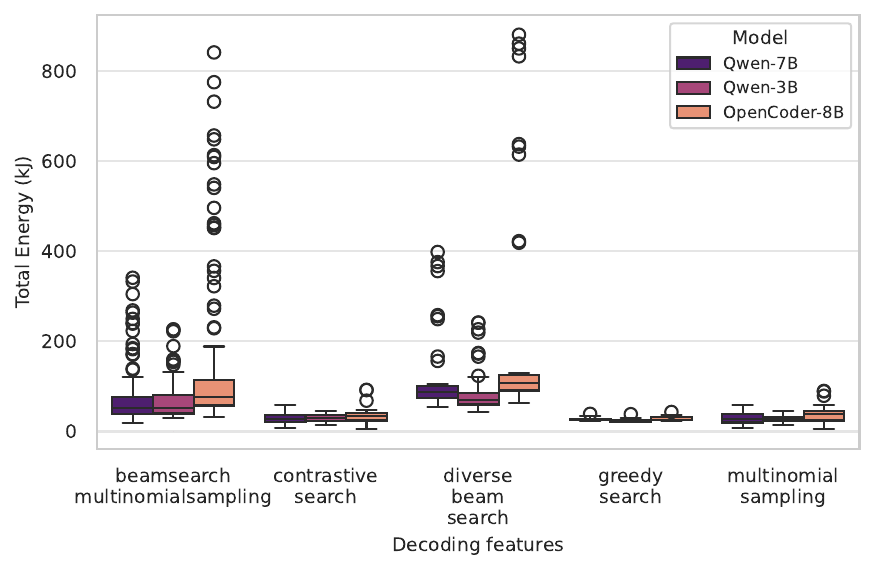}
    \caption{Feature-wise analysis of energy consumption per decoding implementation.}
    \label{fig:feature-wise-generation_total_energy_kj_mean_decoding}
  \end{subfigure}
  \hfill
  \begin{subfigure}{0.43\linewidth}
    \centering
    \includegraphics[width=\linewidth]{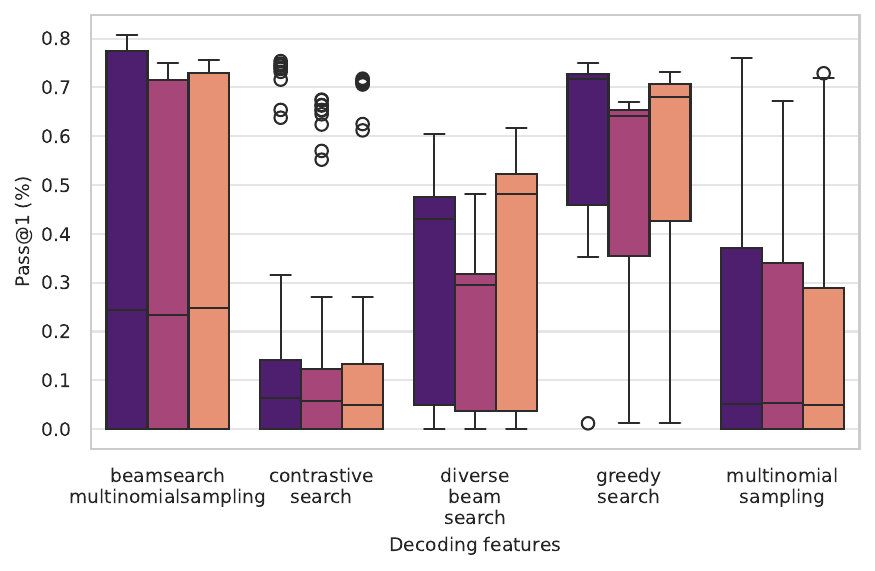}
    \caption{Feature-wise analysis of pass@1 per decoding implementation.}
    \label{fig:feature-wise-pass_1}
  \end{subfigure}
  \caption{Feature-wise analysis of energy consumption and accuracy across decoding implementations.}
  \label{fig:feature-wise-combined}
\end{figure*}

Regarding accuracy, hyperparameters related to repetition control, decoding strategy, and memory usage dominate. 
The no-repeat-ngram parameter prevents the model from generating repeated sequences of n consecutive tokens. 
Setting it to 0 allows the most freedom and is efficient for accuracy in our use-case. 
The low-memory parameter reduces memory usage during generation by recomputing certain operations instead of keeping them in GPU memory. 
While this saves resources, its activation can slightly degrade the quality of token predictions and therefore impact negatively the accuracy. 
Finally, the decoding strategy, which determines how the next token is selected, is also present among the most impactful feature on accuracy. 
Greedy decoding, which always selects the token with the highest predicted probability at each step, improves accuracy in our experiments. 
Figure~\ref{fig:feature-wise-pass_1} confirms this observation, showing that \texttt{greedy search} decoding achieves the highest accuracy among the tested strategies, \texttt{beam search multinomial sampling}, \texttt{contrastive search}, \texttt{diverse beam search} and \texttt{multinomial sampling}. 
At the same time, it is also the least energy-intensive decoding implementation according to Figure~\ref{fig:feature-wise-generation_total_energy_kj_mean_decoding}, making it the most efficient decoding strategy overall. 

\textbf{\textit{Pairwise Analysis}}.
To analyze interactions between generation hyperparameters, we apply the second method proposed by \etal{Guégain}~\cite{guegain2021reducing}, referred to as \emph{pairwise analysis}.
Unlike feature-wise analysis, which considers features independently, this method captures the combined effect of feature pairs on a given metric.
Concretely, a matrix is constructed with one column per valid pair of features. 
For each sampled configuration, the measured metric value is assigned to all columns representing pairs of features included in that configuration.
While the feature-wise analysis identified several hyperparameters with individual effects—such as dynamic caching and greedy decoding—the pairwise analysis allows us to assess whether these effects depend on specific combinations with other parameters or can be further amplified through interactions.
To illustrate this, we focus on configurations using \texttt{greedy search} decoding\footnote{Similar analysis can be performed for any other feature using the replication package described in \citesec{setup}.}.
We fix the decoding strategy to greedy search and report, in \citetable{feature_pair}, the top five feature combinations, whose values differ from the default configuration, and exhibit the strongest impact on energy consumption, latency, and accuracy. 

For energy consumption and latency, we again observe nearly identical rankings, reinforcing the strong correlation between these two metrics.
Combinations of \texttt{greedy search} decoding with hyperparameters such as \texttt{cache}, \texttt{encoder-repetition-penalty}, \texttt{do\allowbreak\_sample}, \texttt{top-k}, and \texttt{top-p} show the strongest impact. 
For instance, a static cache increases the energy consumption by $13.76$~kJ and latency by over $48.12$, while enabling \texttt{do\_sample} reduces them by $8.59$~kJ and $21.79$~s, respectively.
Regarding accuracy, the most significant effects arise from interactions between \texttt{greedy search} and parameters such as \texttt{repetition-penalty}, \texttt{do-sample}, \texttt{top-p}, and \texttt{temperature}. For example, setting \texttt{repetition-penalty} to $0.6$ decline accuracy by $0.65$, and fixing \texttt{top-p} to $0.8$ improves Pass@1 by $0.23$. 
Notably, some parameter values—such as \texttt{top-p} set to $0.8$—appear among the most influential combinations for latency and accuracy, suggesting configurations that offer favorable trade-offs across all metrics.

\begin{result}
\textbf{RQ1:} 
By leveraging variability modeling, we systematically identified how inference hyperparameters influence energy consumption, latency, and accuracy, both individually and through interactions.
In our case study, energy consumption and latency are strongly correlated and are primarily driven by features such as offloaded cache and hyperparameters that increase the number of forward passes and the diversity of hypotheses considered during generation. 
Accuracy, in contrast, is mainly affected by features related to decoding strategy, repetition control, and memory optimization.
While several efficient features correspond to the default settings of Hugging Face Transformers—such as dynamic caching and greedy decoding—our results show that these defaults can be further improved when combined with carefully chosen complementary parameters.
\end{result}

\paragraph*{RQ2: Which Pareto-optimal trade-offs between energy consumption, performance, and accuracy can be identified?}

\begin{figure}[h]
    \centering
    \includegraphics[width=\linewidth]{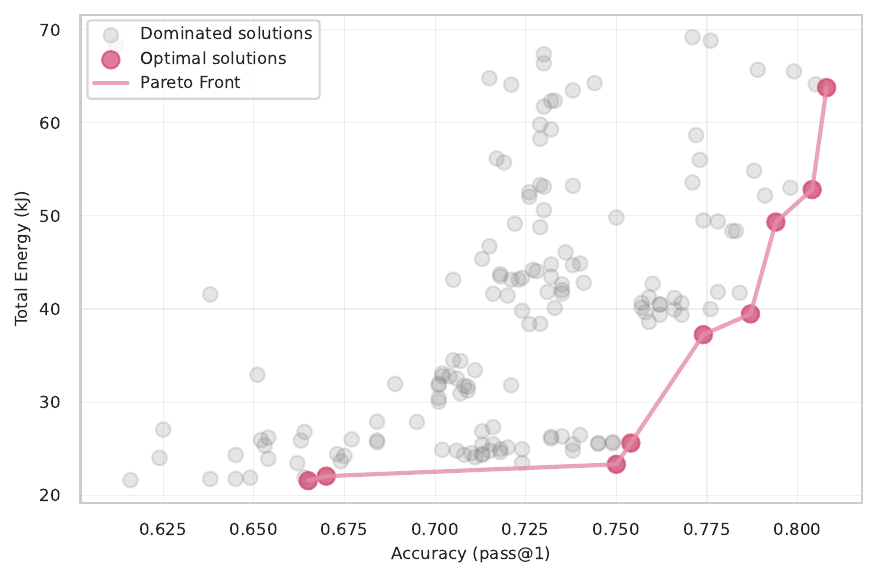}
    \caption{Pareto front illustrating the trade-offs between energy consumption and accuracy across the different configurations and selected \glspl{llm}.}
    \label{fig:pareto-front}
\end{figure}

To analyze trade-offs between these measurements, we computed the Pareto front over the empirically measured configurations for all the selected \glspl{llm}, as shown in Figure \ref{fig:pareto-front}.
It represents configurations that cannot be improved on one metric without degrading another, thereby identifying the most efficient trade-offs in the configuration space.
We focused on energy consumption and accuracy, since latency was shown in RQ1 to be strongly correlated with energy.
Two configurations with a Pass@1 score of zero were excluded, as well as configurations exceeding 70 kJ, since they do not correspond to practically usable inference settings, and their removal improves the clarity of the figure.

The results show that accuracy improvements exhibit diminishing returns as energy consumption increases. 
In the initial low-energy phase (approximately $20$-$25$ kJ), small increases in energy lead to large gains in accuracy, with Pass@1 improving from $0.66$ (with \emph{Qwen-3B}) to $0.75$ (with \emph{Qwen-7B}).
For \emph{Qwen-3B}, sampling is not always activated, dynamic caching is used, along with a single beam and beam group, a low temperature of $0.3$, and top-p set to $0.8$ when sampling is enabled.
For \emph{Qwen-7B}, dynamic caching is employed with either greedy or contrastive decoding. When sampling is disabled, default sampler parameters are used, and when it's enabled, temperature is set to $0.1$ and top-p to $0.95$. Configurations using \emph{Qwen-7B} in this phase achieve an increase of $0.09$ in accuracy compared to \emph{Qwen-3B}, while consuming $7$-$10\%$ more energy.

In the balanced phase (around $35$-$40$ kJ), accuracy gains become more moderate, increasing from $0.75$ to $0.77$.
The corresponding configurations consistently use \emph{Qwen-7B} with dynamic caching, beam search multinomial with two beams, sampling enabled, low temperatures ($0.1$--$0.3$), low top-k values ($5$--$25$), and top-p values of $0.8$--$0.85$.

Finally, in the high-accuracy phase ($49$-$64$ kJ), Pass@1 reaches a maximum of $0.80$, with energy consumption increasing by up to $66\%$ compared to the balanced phase.
These configurations employ \emph{Qwen-7B} with dynamic, static, or hybrid caching, more aggressive decoding strategies such as beam search with three to four beams, temperatures ranging from $0.1$ to $0.7$, and top-p from $0.8$ to $1.0$.

\begin{result}
\textbf{RQ2:}
By combining feature modeling, systematic exploration, and Pareto analysis, we identify distinct classes of inference configurations that balance trade-offs between energy, latency, and accuracy. 
Most accuracy gains can be achieved in a low-energy regime using dynamic caching, greedy or contrastive decoding, a single beam, and default sampling parameters, or a combination of low temperature and high top-p.
Balanced configurations increase energy consumption by introducing two-beam search, sampling with low temperature, smaller top-k, and high top-p values.
Achieving the highest accuracy requires much more energy and more complex configurations, with dynamic, hybrid or static cache, $3$–$4$ beams, low temperature, and high top-p. 
\end{result}

\begin{table*}[t]
\caption{Optimal configurations for each objective and their impact on energy and accuracy. Only a subset of the evaluated parameters is shown for readability.}
\label{table:pareto-objectives}
\renewcommand{\arraystretch}{1}
\centering
 \resizebox{\textwidth}{!}{ 
\begin{tabular}{l|c|c|c|c|c|c|c}
\toprule
\textbf{Objective} &
\multicolumn{3}{c|}{\textbf{Energy-efficient}} &
\textbf{Balanced} &
\multicolumn{3}{c}{\textbf{High accuracy}} \\
\midrule
Model & Qwen-3B & Qwen-7B & Qwen-7B & Qwen-7B & Qwen-7B & Qwen-7B & Qwen-7B\\
Cache & Dynamic & Dynamic & Dynamic & Dynamic & Dynamic & Static & Hybrid \\
Decoding & Greedy & Greedy & Contrastive & Beam Search Multi & Beam Search Multi & Beam Search Multi & Beam Search Multi \\
Number of beams & 1 & 1 & 1 & 2 & 3 & 3 & 4 \\
Beam groups & 1 & 1 & 1 & 1 & 1 & 1 & 1\\
Sampling & Partial & $\times$ & \checkmark & \checkmark & \checkmark & \checkmark & \checkmark \\
Temperature & 0.3 / 1.0 & 1.0 & 0.1 & 0.1 / 0.3 & 0.3 & 0.1 & 0.7\\
Top-k & 50 & 50 & 50 & 5 / 25 & 50 & 50 & 5\\
Top-p & 0.8 / 1.0 & 1.0 & 0.95 & 0.8 / 0.85 & 0.8 & 0.8 & 1.0\\
\midrule
Avg accuracy & 0.66 & 0.75 (+0.09)& 0.75 (+0.00) & 0.77 (+0.02) & 0.79 (+0.02) & 0.80 (+0.01) & 0.80 (+0.00)\\
Avg energy (kJ) & 21.79 & 23.31 (+7\%) & 25.59 (+10\%) & 38.35 (+29\%) & 49.33 (+28.63) & 52.82 (+7\%) & 63.77 (+20.73\%) \\
\bottomrule
\end{tabular}
}
\end{table*}

\paragraph*{RQ3: To what extent can machine learning models accurately predict energy consumption, performance, and accuracy for unseen \gls{llm} inference configurations?}

\citetable{modelsPerf} reports the prediction performance of our models for energy consumption, latency, and accuracy (Pass@1) across different \glspl{llm} and sampling strategies. The \gls{mape} is not reported for accuracy models due to the presence of null values, which are incompatible with \gls{mape}.

Overall, the results indicate that prediction of unseen inference configurations is achievable. However, predictive performance strongly depends on both the target metric and the sampling strategy used to construct the training set. Models trained on configurations generated by all three samplers (\emph{ALL}) consistently achieve the best results across all metrics. They achieve $R^2$ values of $0.95$ for energy, $0.94$ for latency, and $0.99$ for accuracy, along with the lowest errors: for energy, MAE $10.05$ and MAPE $0.16$; for latency, MAE $23.78$ and MAPE $0.17$; and for accuracy, MAE $0.01$. This demonstrates the benefit of combining diverse sampling strategies when constructing the training set, as it provides richer and more representative training data.

Considering individual samplers, performance varies depending on the metric. The 2-wise sampler (\emph{ICPL}) consistently underperforms, yielding the highest errors and lowest $R^2$ values: for energy, MAE $19.59$ and $R^2$ $0.74$; for latency, MAE $37.77$ and $R^2$ $0.78$. Concerning accuracy, the results are almost identical. This suggests that relying solely on ICPL-generated configurations is insufficient to capture the variability of inference behavior.

In contrast, \emph{YASA} achieves competitive performance across all metrics. For energy and latency, it outperforms \emph{RANDOM} in terms of $R^2$, while for accuracy, it reaches the best $R^2$ ($0.01$) and lowest \gls{mae} ($0.99$), tied with the \emph{ALL} sampler. Interestingly, \emph{RANDOM} sampling is also relatively effective compared to other samplers, indicating that even simple or uniform sampling strategies can provide informative training data.
\begin{result}
\textbf{RQ3:}
Overall, the results show that machine learning models can predict energy consumption, latency, and accuracy for unseen \gls{llm} inference configurations when trained on representative samples of the configuration space. Prediction quality, however, depends on both the target metric and the sampling strategy. Consistent with prior work showing that a single well-chosen sample can yield sufficiently accurate models~\cite{kumara2022focloud}, our results confirm that limited data can be effective. At the same time, the models benefit from increased diversity in the training data, making the combination of multiple sampling strategies particularly effective.
\end{result}

%% file: chapters/threats-to-validity.tex
\section{Discussion \& Limitations}
\label{sec:limitations}
In this section, we discuss some implications of our approach and outline the limitations of our study.
\subsection{Practical Usage \& Implications}
Our results show how variability modeling, combined with empirical measurements and predictive models, can serve as a foundation for managing the complexity of \gls{llm} inference configuration. 
By explicitly structuring the configuration space and encoding valid options and interactions, feature models enable systematic exploration while preventing invalid or unsupported configurations. 
They offer a clear advantage over ad hoc or trial-and-error tuning approaches, particularly in complex inference settings with numerous interacting hyperparameters.
Beyond analysis, our approach lays the foundation for increased automation in inference configuration. 
By combining variability modeling with empirical analysis and learned predictors, it becomes possible to automatically recommend inference configurations tailored to user-specific objectives, such as minimizing energy consumption, maximizing performance, or maintaining a target level of accuracy. 
In practice, this could support configuration assistants that recommend efficient inference setups, guiding users toward valid and efficient configurations while abstracting away low-level parameter interactions.

Inference frameworks expose heterogeneous configuration spaces, with options, parameter names, and supported features differing across inference servers. This heterogeneity suggests using inference server–specific feature models that capture platform-dependent variability, rather than relying on a single feature model.
These models could be connected through a higher-level abstraction, such as an abstract feature model supported by a domain-specific language, providing a unified interface for configuring inference across platforms. 
Such an abstraction would allow  platform-specific configurations to be automatically generated and optimized according to user objectives.

\begin{table*}[t!]
\caption{Overall prediction performance across energy, latency, and accuracy. Best results are shown in bold.}
\label{table:modelsPerf}
\renewcommand{\arraystretch}{1}
\begin{subtable}[t]{0.30\linewidth}
\centering
\caption{Prediction performance for energy (kJ).}
\begin{tabular}{l|ccc}
\toprule
\textbf{Sampler} & \textbf{MAE} & $\mathbf{R^2}$ & \textbf{MAPE}\\
\midrule
YASA   & 12.23 & 0.94 & 0.25\\
ICPL   & 19.59        & 0.74 & 0.25\\
RANDOM & 15.94         & 0.91 & 0.30 \\
ALL    & \best{10.05}  & \best{0.95} & \best{0.16} \\
\bottomrule
\end{tabular}
\end{subtable}
\hfill
\begin{subtable}[t]{0.30\linewidth}
\centering
\caption{Prediction performance for latency (s).}
\begin{tabular}{l|ccc}
\toprule
\textbf{Sampler} & \textbf{MAE} & $\mathbf{R^2}$ & \textbf{MAPE}\\
\midrule
YASA   & 27.72         & 0.93 & 0.25\\
ICPL   & 37.77         & 0.78 & 0.23\\
RANDOM & 32.99 & 0.91 & 0.32 \\
ALL    & \best{23.78}  & \best{0.94} & \best{0.17}\\
\bottomrule
\end{tabular}
\end{subtable}
\hfill
\begin{subtable}[t]{0.30\linewidth}
\centering
\caption{Prediction performance for pass@1.}
\begin{tabular}{l|cc}
\toprule
\textbf{Sampler} & \textbf{MAE} & $\mathbf{R^2}$ \\
\midrule
YASA   & \best{0.01}          & \best{0.99} \\
ICPL   & 0.02  & 0.97 \\
RANDOM & 0.02          & 0.96 \\
ALL    & \best{0.01}  & \best{0.99} \\
\bottomrule
\end{tabular}
\end{subtable}

\end{table*}

\subsection{Limits \& Threats to Validity}

While our study provides valuable insights into leveraging variability modeling for the efficient and sustainable configuration of \gls{llm}, several limitations and threats to validity must be considered.

\textbf{\textit{External validity.}}
Our evaluation relies on a single dataset, \textit{HumanEval+}, which focuses on Python code generation tasks. As a result, it does not capture the diversity of inference behaviors observed in other tasks, such as summarization and translation, nor across other programming languages. 
Consequently, the relationship between configuration parameters and metrics such as energy consumption, latency, and accuracy may differ across tasks and datasets, limiting the external validity of our findings. 
Similarly, all experiments were conducted using a single inference server, Hugging Face Transformers.
Although this framework offers fine-grained control over generation parameters, it is not representative of all inference servers, which may expose different configuration options or implement similar parameters differently. 
Extending the analysis to other inference platforms could therefore reveal additional constraints and interactions that may influence energy consumption, latency, and accuracy. 
In addition, our study considered only three \glspl{llm}. 
Given the rapid evolution of \gls{llm} architectures, training techniques, and optimization methods, the observed results may not generalize to other models or future systems.

Despite these limitations, the proposed approach is generic and can be applied to other datasets,  inference servers, and \glspl{llm}, provided that an appropriate variability model is defined.

\textbf{\textit{Construct validity.}}
The proposed feature model has inherent limitations. 
Due to the large number of available hyperparameters, we selected only those relevant to our use case, and continuous parameters were discretized into a limited set of representative values. 
While this was necessary to keep the configuration space tractable, it may hide finer-grained effects or overlook configurations that could yield different trade-offs.

Moreover, some dependencies and constraints between parameters may not be fully captured by the feature model.
This is reflected by the fact that approximately $3\%$ of the generated configurations could not be executed successfully, indicating incomplete or missing constraints.

\textbf{\textit{Internal validity and reproducibility.}}
All measurements were conducted on NVIDIA A100 GPUs in a controlled environment. While this experimental setup ensures consistency and reproducibility, it does not capture the full variability of hardware configurations and deployment environments encountered in practice, which can significantly influence inference behavior.

\textbf{Feature modeling challenges.}
\gls{ai} and \gls{llm} ecosystems evolve rapidly, with new architectures, quantization techniques, and serving frameworks emerging frequently. 
As a result, feature models require continuous maintenance to remain aligned with this evolution, which can hinder reproducibility and long-term generalization.
Beyond generation hyperparameters, other factors such as hardware configuration and deployment environment also influence inference behavior. 
Including these additional sources of variability would increase the complexity of the variability model and complicate both its construction and maintenance. 
In particular, hardware-related parameters can make constraint generation more difficult, and failures—such as out-of-memory errors—are harder to interpret, as they may arise from multiple interacting factors.
These challenges, however, are well-known in the variability modeling community. 
Prior work has explored the use of machine learning techniques for automatic constraint inference~\cite{temple2017learning}, and several studies have investigated strategies for managing evolving variability models~\cite{marques2019software}.

%% file: chapters/conclusion.tex
\section{Conclusion \& Future Work}
\label{sec:conclusion}
Variability modeling techniques have long been used to manage and reason about configurable software systems. 

In this paper, we extend this perspective to \glspl{llm}, viewing their inference process as a highly configurable system characterized by numerous interdependent hyperparameters. 
By representing these configurations through feature modeling, we bridge the gap between the variability management and machine learning communities.
Our results show that variability modeling provides a systematic framework for exploring the complex configuration space of \glspl{llm}. It enables structured analysis of how individual and combined parameters affect metrics such as energy consumption, performance, accuracy, and supports the construction of predictive models capable of estimating these metrics for unseen configurations. 
This approach directly addresses the combinatorial explosion inherent to exhaustive configuration exploration, allowing trade-offs to be evaluated more efficiently.

Looking forward, we plan to extend this work by:
\emph{(i)} enriching the variability model to incorporate additional variability level, such as hardware configurations and deployment options;
\emph{(ii)} leveraging our approach for adaptive reconfiguration during inference, enabling runtime trade-off management; and
\emph{(iii)} extending the methodology to other inference servers, datasets, and models.

%% file: chapters/acknowledgments.tex
\label{sec:acknowledgments}
\begin{acks}
We thank the anonymous reviewers for their insightful comments and feedback. This work received funding from the France 2030 program, managed by the French National Research Agency under grant agreement No. ANR-23-PECL-0003.
\end{acks}

%% file: bibfile.bib
@inproceedings{guegain2021reducing,
  title={On reducing the energy consumption of software product lines},
  author={Gu{\'e}gain, {\'E}douard and Quinton, Cl{\'e}ment and Rouvoy, Romain},
  booktitle={Proceedings of the 25th ACM International Systems and Software Product Line Conference-Volume A},
  pages={89--99},
  year={2021}
}

@inproceedings{couto2017products,
  title={Products go green: Worst-case energy consumption in software product lines},
  author={Couto, Marco and Borba, Paulo and Cunha, J{\'a}come and Fernandes, Jo{\~a}o Paulo and Pereira, Rui and Saraiva, Jo{\~a}o},
  booktitle={Proceedings of the 21st International Systems and Software Product Line Conference-Volume A},
  pages={84--93},
  year={2017}
}

@inproceedings{guegain2025exploring,
  title={Exploring Performance of Configurable Software Systems: the JHipster Case Study},
  author={Gu{\'e}gain, {\'E}douard and Bonvoisin, Alexandre and Acher, Mathieu and Quinton, Cl{\'e}ment and Rouvoy, Romain},
  booktitle={EASE'25-29th International Conference on Evaluation and Assessment in Software Engineering},
  year={2025}
}

@inproceedings{kwon2023efficient,
  title={Efficient Memory Management for Large Language Model Serving with PagedAttention},
  author={Woosuk Kwon and Zhuohan Li and Siyuan Zhuang and Ying Sheng and Lianmin Zheng and Cody Hao Yu and Joseph E. Gonzalez and Hao Zhang and Ion Stoica},
  booktitle={Proceedings of the ACM SIGOPS 29th Symposium on Operating Systems Principles},
  year={2023}
}

@article{pereira2021learning,
  title={Learning software configuration spaces: A systematic literature review},
  author={Pereira, Juliana Alves and Acher, Mathieu and Martin, Hugo and J{\'e}z{\'e}quel, Jean-Marc and Botterweck, Goetz and Ventresque, Anthony},
  journal={Journal of Systems and Software},
  volume={182},
  pages={111044},
  year={2021},
  publisher={Elsevier}
}

@article{galindo2019automated,
  title={Automated analysis of feature models: Quo vadis?},
  author={Galindo, Jos{\'e} A and Benavides, David and Trinidad, Pablo and Guti{\'e}rrez-Fern{\'a}ndez, Antonio-Manuel and Ruiz-Cort{\'e}s, Antonio},
  journal={Computing},
  volume={101},
  number={5},
  pages={387--433},
  year={2019},
  publisher={Springer}
}

@article{quinton2016saloon,
  title={SALOON: a platform for selecting and configuring cloud environments},
  author={Quinton, Cl{\'e}ment and Romero, Daniel and Duchien, Laurence},
  journal={Software: Practice and Experience},
  volume={46},
  number={1},
  pages={55--78},
  year={2016},
  publisher={Wiley Online Library}
}

@article{kumara2022focloud,
  title={FOCloud: feature model guided performance prediction and explanation for deployment configurable cloud applications},
  author={Kumara, Indika and Ariz, Mohamed Hameez and Chhetri, Mohan Baruwal and Mohammadi, Majid and Van Den Heuvel, Willem-Jan and Tamburri, Damian A},
  journal={IEEE Transactions on Services Computing},
  volume={16},
  number={1},
  pages={302--314},
  year={2022},
  publisher={IEEE}
}

@article{guo2018data,
  title={Data-efficient performance learning for configurable systems},
  author={Guo, Jianmei and Yang, Dingyu and Siegmund, Norbert and Apel, Sven and Sarkar, Atrisha and Valov, Pavel and Czarnecki, Krzysztof and Wasowski, Andrzej and Yu, Huiqun},
  journal={Empirical Software Engineering},
  volume={23},
  number={3},
  pages={1826--1867},
  year={2018},
  publisher={Springer}
}

@article{trujillo2020multiple,
  title={Multiple software product lines to configure applications of internet of things},
  author={Trujillo-Tzanahua, Guadalupe-Isaura and Ju{\'a}rez-Mart{\'\i}nez, Ulises and Aguilar-Lasserre, Alberto-Alfonso and Cort{\'e}s-Verd{\'\i}n, Mar{\'\i}a-Karen and Azzaro-Pantel, Catherine},
  journal={IET Software},
  volume={14},
  number={2},
  pages={165--175},
  year={2020},
  publisher={Wiley Online Library}
}

@article{brugali2020software,
  title={Software product line engineering for robotics},
  author={Brugali, Davide},
  journal={Software Engineering for Robotics},
  pages={1--28},
  year={2020},
  publisher={Springer}
}

@inproceedings{gomez2024exploring,
  title={Exploring the use of software product lines for the combination of machine learning models},
  author={Gomez-Vazquez, Marcos and Cabot, Jordi},
  booktitle={Proceedings of the 28th ACM International Systems and Software Product Line Conference},
  pages={26--29},
  year={2024}
}

@inproceedings{ghofrani2019applying,
  title={Applying product line engineering concepts to deep neural networks},
  author={Ghofrani, Javad and Kozegar, Ehsan and Fehlhaber, Anna Lena and Soorati, Mohammad Divband},
  booktitle={Proceedings of the 23rd International Systems and Software Product Line Conference-Volume A},
  pages={72--77},
  year={2019}
}

@inproceedings{camillieri2016towards,
  title={Towards a software product line for machine learning workflows: Focus on supporting evolution},
  author={Camillieri, C{\'e}cile and Parisi, Luca and Blay-Fornarino, Mireille and Precioso, Fr{\'e}d{\'e}ric and Riveill, Michel and Cancela-Vaz, Jo{\"e}l},
  booktitle={10th Workshop on Models and Evolution co-located with ACM/IEEE 19th International Conference on Model Driven Engineering Languages and Systems (MODELS 2016)},
  year={2016}
}

@article{fernandez2025energy,
  title={Energy considerations of large language model inference and efficiency optimizations},
  author={Fernandez, Jared and Na, Clara and Tiwari, Vashisth and Bisk, Yonatan and Luccioni, Sasha and Strubell, Emma},
  journal={arXiv preprint arXiv:2504.17674},
  year={2025}
}

@article{wu2025unveiling,
  title={Unveiling environmental impacts of large language model serving: A functional unit view},
  author={Wu, Yanran and Hua, Inez and Ding, Yi},
  journal={arXiv preprint arXiv:2502.11256},
  year={2025}
}

@inproceedings{patel2024characterizing,
  title={Characterizing power management opportunities for llms in the cloud},
  author={Patel, Pratyush and Choukse, Esha and Zhang, Chaojie and Goiri, {\'I}{\~n}igo and Warrier, Brijesh and Mahalingam, Nithish and Bianchini, Ricardo},
  booktitle={Proceedings of the 29th ACM International Conference on Architectural Support for Programming Languages and Operating Systems, Volume 3},
  pages={207--222},
  year={2024}
}

@article{stojkovic2024towards,
  title={Towards greener llms: Bringing energy-efficiency to the forefront of llm inference},
  author={Stojkovic, Jovan and Choukse, Esha and Zhang, Chaojie and Goiri, Inigo and Torrellas, Josep},
  journal={arXiv preprint arXiv:2403.20306},
  year={2024}
}

@article{wilkins2024offline,
  title={Offline energy-optimal llm serving: Workload-based energy models for llm inference on heterogeneous systems},
  author={Wilkins, Grant and Keshav, Srinivasan and Mortier, Richard},
  journal={ACM SIGENERGY Energy Informatics Review},
  volume={4},
  number={5},
  pages={113--119},
  year={2024},
  publisher={ACM New York, NY, USA}
}

@inproceedings{luccioni2024power,
  title={Power hungry processing: Watts driving the cost of AI deployment?},
  author={Luccioni, Sasha and Jernite, Yacine and Strubell, Emma},
  booktitle={Proceedings of the 2024 ACM conference on fairness, accountability, and transparency},
  pages={85--99},
  year={2024}
}

@article{maliakel2025investigating,
  title={Investigating Energy Efficiency and Performance Trade-offs in LLM Inference Across Tasks and DVFS Settings},
  author={Maliakel, Paul Joe and Ilager, Shashikant and Brandic, Ivona},
  journal={arXiv preprint arXiv:2501.08219},
  year={2025}
}

@inproceedings{stojkovic2025dynamollm,
  title={Dynamollm: Designing llm inference clusters for performance and energy efficiency},
  author={Stojkovic, Jovan and Zhang, Chaojie and Goiri, {\'I}{\~n}igo and Torrellas, Josep and Choukse, Esha},
  booktitle={2025 IEEE International Symposium on High Performance Computer Architecture (HPCA)},
  pages={1348--1362},
  year={2025},
  organization={IEEE}
}

@article{husom2025sustainable,
  title={Sustainable llm inference for edge ai: Evaluating quantized llms for energy efficiency, output accuracy, and inference latency},
  author={Husom, Erik Johannes and Goknil, Arda and Astekin, Merve and Shar, Lwin Khin and K{\aa}sen, Andre and Sen, Sagar and Mithassel, Benedikt Andreas and Soylu, Ahmet},
  journal={arXiv preprint arXiv:2504.03360},
  year={2025}
}

@inproceedings{samsi2023words,
  title={From words to watts: Benchmarking the energy costs of large language model inference},
  author={Samsi, Siddharth and Zhao, Dan and McDonald, Joseph and Li, Baolin and Michaleas, Adam and Jones, Michael and Bergeron, William and Kepner, Jeremy and Tiwari, Devesh and Gadepally, Vijay},
  booktitle={2023 IEEE High Performance Extreme Computing Conference (HPEC)},
  pages={1--9},
  year={2023},
  organization={IEEE}
}

@article{coignion2024green,
  title={Green My LLM: Studying the key factors affecting the energy consumption of code assistants},
  author={Coignion, Tristan and Quinton, Cl{\'e}ment and Rouvoy, Romain},
  journal={arXiv preprint arXiv:2411.11892},
  year={2024}
}

@inproceedings{lazuka2024llm,
  title={Llm-pilot: Characterize and optimize performance of your llm inference services},
  author={Lazuka, Malgorzata and Anghel, Andreea and Parnell, Thomas},
  booktitle={SC24: International Conference for High Performance Computing, Networking, Storage and Analysis},
  pages={1--18},
  year={2024},
  organization={IEEE}
}

@article{zhao2023survey,
  title={A survey of large language models},
  author={Zhao, Wayne Xin and Zhou, Kun and Li, Junyi and Tang, Tianyi and Wang, Xiaolei and Hou, Yupeng and Min, Yingqian and Zhang, Beichen and Zhang, Junjie and Dong, Zican and others},
  journal={arXiv preprint arXiv:2303.18223},
  volume={1},
  number={2},
  year={2023}
}

@inproceedings{johansen2012algorithm,
  title={An algorithm for generating t-wise covering arrays from large feature models},
  author={Johansen, Martin Fagereng and Haugen, {\O}ystein and Fleurey, Franck},
  booktitle={Proceedings of the 16th International Software Product Line Conference-Volume 1},
  pages={46--55},
  year={2012}
}

@inproceedings{martinez2025impact,
  title={The impact of hyperparameters on large language model inference performance: An evaluation of vllm and huggingface pipelines},
  author={Martinez, Matias},
  booktitle={Proceedings of the 33rd ACM International Conference on the Foundations of Software Engineering},
  pages={1672--1678},
  year={2025}
}

@article{jagannadharao2024beginner,
  title={A Beginner's Guide to Power and Energy Measurement and Estimation for Computing and Machine Learning},
  author={Jagannadharao, Akshaya and Beckage, Nicole and Biswas, Sovan and Egan, Hilary and Gafur, Jamil and Metsch, Thijs and Nafus, Dawn and Raffa, Giuseppe and Tripp, Charles},
  journal={arXiv preprint arXiv:2412.17830},
  year={2024}
}

@article{fu2025llmco2,
  title={Llmco2: Advancing accurate carbon footprint prediction for llm inferences},
  author={Fu, Zhenxiao and Chen, Fan and Zhou, Shan and Li, Haitong and Jiang, Lei},
  journal={ACM SIGENERGY Energy Informatics Review},
  volume={5},
  number={2},
  pages={63--68},
  year={2025},
  publisher={ACM New York, NY, USA}
}

@phdthesis{temple2017learning,
  title={Learning-based performance specialization of configurable systems},
  author={Temple, Paul and Acher, Mathieu and J{\'e}z{\'e}quel, Jean-Marc and Noel-Baron, L{\'e}o and Galindo, Jos{\'e} A},
  year={2017},
  school={IRISA, Inria Rennes; University of Rennes 1}
}

@article{chen2021evaluating,
  title={Evaluating large language models trained on code},
  author={Chen, Mark and Tworek, Jerry and Jun, Heewoo and Yuan, Qiming and Pinto, Henrique Ponde De Oliveira and Kaplan, Jared and Edwards, Harri and Burda, Yuri and Joseph, Nicholas and Brockman, Greg and others},
  journal={arXiv preprint arXiv:2107.03374},
  year={2021}
}

@article{liu2023your,
  title={Is your code generated by chatgpt really correct? rigorous evaluation of large language models for code generation},
  author={Liu, Jiawei and Xia, Chunqiu Steven and Wang, Yuyao and Zhang, Lingming},
  journal={Advances in Neural Information Processing Systems},
  volume={36},
  pages={21558--21572},
  year={2023}
}

@inproceedings{coignion2025faster,
  title={When Faster Isn't Greener: The Hidden Costs of LLM-Based Code Optimization},
  author={Coignion, Tristan and Quinton, Cl{\'e}ment and Rouvoy, Romain},
  booktitle={ASE'25-40th International Conference on Automated Software Engineering},
  year={2025}
}

@article{chen2009variability,
  title={Variability management in software product lines: a systematic review},
  author={Chen, Lianping and Ali Babar, Muhammad and Ali, Nour},
  year={2009},
  publisher={University of Limerick}
}

@inproceedings{czarnecki2012cool,
  title={Cool features and tough decisions: a comparison of variability modeling approaches},
  author={Czarnecki, Krzysztof and Gr{\"u}nbacher, Paul and Rabiser, Rick and Schmid, Klaus and W{a}sowski, Andrzej},
  booktitle={Proceedings of the 6th International workshop on variability modeling of software-intensive systems},
  pages={173--182},
  year={2012}
}

@article{schaefer2012software,
  title={Software diversity: state of the art and perspectives},
  author={Schaefer, Ina and Rabiser, Rick and Clarke, Dave and Bettini, Lorenzo and Benavides, David and Botterweck, Goetz and Pathak, Animesh and Trujillo, Salvador and Villela, Karina},
  journal={International Journal on Software Tools for Technology Transfer},
  volume={14},
  pages={477--495},
  year={2012},
  publisher={Springer}
}

@inproceedings{berger2013survey,
  title={A survey of variability modeling in industrial practice},
  author={Berger, Thorsten and Rublack, Ralf and Nair, Divya and Atlee, Joanne M and Becker, Martin and Czarnecki, Krzysztof and W{a}sowski, Andrzej},
  booktitle={Proceedings of the 7th International Workshop on Variability Modelling of Software-intensive Systems},
  pages={1--8},
  year={2013}
}

@inproceedings{czarnecki2007feature,
  title={Feature diagrams and logics: There and back again},
  author={Czarnecki, Krzysztof and Wasowski, Andrzej},
  booktitle={11th International Software Product Line Conference (SPLC 2007)},
  pages={23--34},
  year={2007},
  organization={IEEE}
}

@inproceedings{batory2005feature,
  title={Feature models, grammars, and propositional formulas},
  author={Batory, Don},
  booktitle={International Conference on Software Product Lines},
  pages={7--20},
  year={2005},
  organization={Springer}
}

@inproceedings{alves2020sampling,
  title={Sampling effect on performance prediction of configurable systems: A case study},
  author={Alves Pereira, Juliana and Acher, Mathieu and Martin, Hugo and J{\'e}z{\'e}quel, Jean-Marc},
  booktitle={Proceedings of the ACM/SPEC International Conference on Performance Engineering},
  pages={277--288},
  year={2020}
}

@article{agh2024software,
  title={Software product line testing: a systematic literature review},
  author={Agh, Halimeh and Azamnouri, Aidin and Wagner, Stefan},
  journal={Empirical Software Engineering},
  volume={29},
  number={6},
  pages={146},
  year={2024},
  publisher={Springer}
}

@inproceedings{acher2023generative,
  title={Generative AI for reengineering variants into software product lines: an experience report},
  author={Acher, Mathieu and Martinez, Jabier},
  booktitle={Proceedings of the 27th International Systems and Software Product Line Conference-Vol. B},
  pages={57--66},
  year={2023}
}

@inproceedings{stumpfle2024automating,
  title={Automating Software Product Line Adoption Based on Feature Models Using Large Language Models},
  author={St{\"u}mpfle, Johannes and Baum, Sebastian and Dittler, Daniel and Jazdi, Nasser and Weyrich, Michael},
  booktitle={2024 IEEE 29th International Conference on Emerging Technologies and Factory Automation (ETFA)},
  pages={1--4},
  year={2024},
  organization={IEEE}
}

@inproceedings{acher2023programming,
  title={On programming variability with large language model-based assistant},
  author={Acher, Mathieu and Duarte, Jos{\'e} Galindo and J{\'e}z{\'e}quel, Jean-Marc},
  booktitle={Proceedings of the 27th International Systems and Software Product Line Conference-Vol. A},
  pages={8--14},
  year={2023}
}

@inproceedings{galindo2023large,
  title={Large language models to generate meaningful feature model instances},
  author={Galindo, Jos{\'e} A and Dominguez, Antonio J and White, Jules and Benavides, David},
  booktitle={Proceedings of the 27th ACM International Systems and Software Product Line Conference-Volume A},
  pages={15--26},
  year={2023}
}

@inproceedings{kebaili2024empirical,
  title={An Empirical Study on Leveraging LLMs for Metamodels and Code Co-evolution},
  author={Kebaili, Zohra Kaouter and Khelladi, Djamel Eddine and Acher, Mathieu and Barais, Olivier},
  booktitle={European Conference on Modelling Foundations and Applications (ECMFA 2024)},
  volume={23},
  number={3},
  pages={1--14},
  year={2024},
  organization={Journal of Object Technology}
}

@inproceedings{zine2025llm,
  title={{LLM-based Co-Evolution of Configurable Software Systems}},
  author={Zine, Nada and Quinton, Cl{\'e}ment and Rouvoy, Romain},
  booktitle={Proceedings of the 2025 29th ACM International Systems and Software Product Line Conference-Volume A},
  pages={27--38},
  year={2025}
}

@inproceedings{lopez2025configuration,
  title={Configuration Bugs Classification using LLMs and Encoders},
  author={Lopez-Duran, Noelia and Romero Organv{\'\i}dez, David and Cruz, Ferm{\'\i}n and Benavides, David},
  booktitle={Proceedings of the 2025 29th ACM International Systems and Software Product Line Conference-Volume A},
  pages={190--200},
  year={2025}
}

@article{wolf2019huggingface,
  title={Huggingface's transformers: State-of-the-art natural language processing},
  author={Wolf, Thomas and Debut, Lysandre and Sanh, Victor and Chaumond, Julien and Delangue, Clement and Moi, Anthony and Cistac, Pierric and Rault, Tim and Louf, R{\'e}mi and Funtowicz, Morgan and others},
  journal={arXiv preprint arXiv:1910.03771},
  year={2019}
}

@article{park2025survey,
  title={A Survey on Inference Engines for Large Language Models: Perspectives on Optimization and Efficiency},
  author={Park, Sihyeong and Jeon, Sungryeol and Lee, Chaelyn and Jeon, Seokhun and Kim, Byung-Soo and Lee, Jemin},
  journal={arXiv preprint arXiv:2505.01658},
  year={2025}
}

@inproceedings{alizadeh2025language,
  title={Language Models in Software Development Tasks: An Experimental Analysis of Energy and Accuracy},
  author={Alizadeh, Negar and Belchev, Boris and Saurabh, Nishant and Kelbert, Patricia and Castor, Fernando},
  booktitle={2025 IEEE/ACM 22nd International Conference on Mining Software Repositories (MSR)},
  pages={725--736},
  year={2025},
  organization={IEEE}
}

@inproceedings{krieter2020yasa,
  title={YASA: yet another sampling algorithm},
  author={Krieter, Sebastian and Th{\"u}m, Thomas and Schulze, Sandro and Saake, Gunter and Leich, Thomas},
  booktitle={Proceedings of the 14th International Working Conference on Variability Modelling of Software-Intensive Systems},
  pages={1--10},
  year={2020}
}

@inproceedings{al2016tool,
  title={Tool demo: testing configurable systems with featureIDE},
  author={Al-Hajjaji, Mustafa and Meinicke, Jens and Krieter, Sebastian and Schr{\"o}ter, Reimar and Th{\"u}m, Thomas and Leich, Thomas and Saake, Gunter},
  booktitle={Proceedings of the 2016 ACM SIGPLAN International Conference on Generative Programming: Concepts and Experiences},
  pages={173--177},
  year={2016}
}

@article{zhuo2024bigcodebench,
  title={Bigcodebench: Benchmarking code generation with diverse function calls and complex instructions},
  author={Zhuo, Terry Yue and Vu, Minh Chien and Chim, Jenny and Hu, Han and Yu, Wenhao and Widyasari, Ratnadira and Yusuf, Imam Nur Bani and Zhan, Haolan and He, Junda and Paul, Indraneil and others},
  journal={Preprint arXiv:2406.15877},
  year={2024}
}

@inproceedings{huyghe2025browserfm,
author = {Huyghe, Maxime and Quinton, Cl\'{e}ment and Rudametkin, Walter},
title = {Taming the Variability of Browser Fingerprints},
year = {2024},
isbn = {9798400705939},
booktitle = {Proceedings of the 28th ACM International Systems and Software Product Line Conference},
pages = {66–71},
numpages = {6},
series = {SPLC '24}
}

@inproceedings{huyghe2025fp,
  title={FP-Rainbow: Fingerprint-Based Browser Configuration Identification},
  author={Huyghe, Maxime and Rudametkin, Walter and Quinton, Cl{\'e}ment},
  booktitle={Proceedings of the ACM on Web Conference 2025},
  pages={4325--4335},
  year={2025}
}

@article{kuiter2025configurable,
  title={How configurable is the Linux Kernel? Analyzing two decades of feature-model history},
  author={Kuiter, Elias and Sundermann, Chico and Th{\"u}m, Thomas and Hess, Tobias and Krieter, Sebastian and Saake, Gunter},
  journal={ACM Transactions on Software Engineering and Methodology},
  year={2025},
  publisher={ACM New York, NY}
}

@article{chicco2021coefficient,
  title={The coefficient of determination R-squared is more informative than SMAPE, MAE, MAPE, MSE and RMSE in regression analysis evaluation},
  author={Chicco, Davide and Warrens, Matthijs J and Jurman, Giuseppe},
  journal={Peerj computer science},
  volume={7},
  pages={e623},
  year={2021},
  publisher={PeerJ Inc.}
}

@article{breiman2001random,
  title={Random forests},
  author={Breiman, Leo},
  journal={Machine learning},
  volume={45},
  number={1},
  pages={5--32},
  year={2001},
  publisher={Springer}
}

@article{zheng2025towards,
  title={Towards an understanding of large language models in software engineering tasks},
  author={Zheng, Zibin and Ning, Kaiwen and Zhong, Qingyuan and Chen, Jiachi and Chen, Wenqing and Guo, Lianghong and Wang, Weicheng and Wang, Yanlin},
  journal={Empirical Software Engineering},
  volume={30},
  number={2},
  pages={50},
  year={2025},
  publisher={Springer}
}

@article{zhou2024survey,
  title={A survey on efficient inference for large language models},
  author={Zhou, Zixuan and Ning, Xuefei and Hong, Ke and Fu, Tianyu and Xu, Jiaming and Li, Shiyao and Lou, Yuming and Wang, Luning and Yuan, Zhihang and Li, Xiuhong and others},
  journal={arXiv preprint arXiv:2404.14294},
  year={2024}
}

@inproceedings{zhao2022green,
  title={A green (er) world for ai},
  author={Zhao, Dan and Frey, Nathan C and McDonald, Joseph and Hubbell, Matthew and Bestor, David and Jones, Michael and Prout, Andrew and Gadepally, Vijay and Samsi, Siddharth},
  booktitle={2022 IEEE International Parallel and Distributed Processing Symposium Workshops (IPDPSW)},
  pages={742--750},
  year={2022},
  organization={IEEE}
}

@article{marques2019software,
  title={Software product line evolution: A systematic literature review},
  author={Marques, Ma{\'\i}ra and Simmonds, Jocelyn and Rossel, Pedro O and Bastarrica, Mar{\'\i}a Cecilia},
  journal={Information and Software Technology},
  volume={105},
  pages={190--208},
  year={2019},
  publisher={Elsevier}
}

@article{nik2025impact,
  title={Impact of decoding strategies on GPU energy usage in large language model text generation},
  author={Nik, Alireza and Riegler, Michael A and Halvorsen, P{\aa}l},
  journal={Scientific Reports},
  year={2025},
  publisher={Nature Publishing Group}
}

@article{shi2402thorough,
  title={A thorough examination of decoding methods in the era of llms, 2024},
  author={Shi, Chufan and Yang, Haoran and Cai, Deng and Zhang, Zhisong and Wang, Yifan and Yang, Yujiu and Lam, Wai},
  journal={URL https://arxiv. org/abs/2402.06925}
}

@inproceedings{arias2025decoding,
  title={Decoding decoded: Understanding hyperparameter effects in open-ended text generation},
  author={Arias, Esteban Garces and Li, Meimingwei and Heumann, Christian and A{\ss}enmacher, Matthias},
  booktitle={Proceedings of the 31st International Conference on Computational Linguistics},
  pages={9992--10020},
  year={2025}
}

@article{caravaca2025prompts,
  title={From Prompts to Power: Measuring the Energy Footprint of LLM Inference},
  author={Caravaca, Francisco and Cuevas, {\'A}ngel and Cuevas, Rub{\'e}n},
  journal={arXiv preprint arXiv:2511.05597},
  year={2025}
}

@inproceedings{stumpfle2025large,
  title={Large language model assisted transformation of software variants into a software product line},
  author={St{\"u}mpfle, Johannes and Atray, Devansh and Jazdi, Nasser and Weyrich, Michael},
  booktitle={2025 IEEE/ACM 22nd International Conference on Software and Systems Reuse (ICSR)},
  pages={12--20},
  year={2025},
  organization={IEEE}
}
